\documentclass[runningheads]{llncs}

\usepackage{graphicx}
\usepackage{amsmath}
\usepackage{amssymb}
\usepackage{mathtools}
\usepackage{amsfonts}
\usepackage{mathtools}
\usepackage{multirow}
\usepackage{units}
\usepackage{diagbox}
\usepackage{csvsimple}
\usepackage{pgfplotstable}
\usepackage{longtable}
\usepackage{float}
\usepackage{caption}
\usepackage{subcaption}
\usepackage{graphicx}
\usepackage{comment}
\usepackage{wrapfig}

\newcommand{\eg}{e.\,g., }
\newcommand{\ie}{i.\,e., }

\newcolumntype{H}{>{\setbox0=\hbox\bgroup}c<{\egroup}@{}}

\begin{document}

\title{Time Series Forecasting Models Copy the Past: How to Mitigate}

\author{Chrysoula Kosma \inst{1} \and
Giannis Nikolentzos\inst{1} \and
Nancy Xu \inst{2}\and
Michalis Vazirgiannis \inst{1, 2}}

\authorrunning{C. Kosma et al.}

\institute{\'Ecole Polytechnique, Palaiseau, France \and
KTH Royal Institute of Technology, Stockholm, Sweden}

\maketitle              

\begin{abstract}
Time series forecasting is at the core of important application domains  posing significant challenges to machine learning algorithms.
Recently neural network architectures have been widely applied to the problem of time series forecasting.
Most of these models are trained by minimizing a loss function that measures predictions' deviation from the real values.
Typical loss functions include  mean squared error (MSE) and mean absolute error (MAE).
In the presence of noise and uncertainty, neural network models tend to replicate the last observed value of the time series, thus limiting their applicability to real-world data.
In this paper, we provide a formal definition of the above problem and we also give some examples of forecasts where the problem is observed.
We also propose a regularization term penalizing the replication of previously seen values.
We evaluate the proposed regularization term both on synthetic and real-world datasets.
Our results indicate that the regularization term mitigates to some extent the aforementioned problem and gives rise to more robust models.

\end{abstract}

\section{Introduction}

Time series are ubiquitous in several application domains including quantitative finance, seismology and meteorology, just to name a few.
Due to this abundance of time series data, the problem of time series forecasting has recently emerged as a very important task with applications ranging from traffic forecasting to financial investment.
Indeed, accurate forecasting is of great importance since it can improve future decisions which is the main objective in a number of scenarios.
For example, traffic forecasting seeks to predict future web traffic to make decisions for better congestion control~\cite{hamed1995short}.
Moreover, forecasting the spread of COVID-19 is of paramount importance to governments and policymakers in order to impose measures to combat the spread of the disease~\cite{chimmula2020time}.

With the advent of deep learning, deep neural networks have become the dominant approach to the problem of time series forecasting.
For instance, models and layers such as Long Short-Term Memory~\cite{hochreiter1997long}, Gated Recurrent Units~\cite{cho2014learning} and Temporal Convolution Networks~\cite{bai2018empirical} have proven to be very successful in temporal modeling.
Specifically, these models have demonstrated great success in capturing complex nonlinear dependencies between variables and time, while they usually operate on raw time series data, thus requiring considerably less human effort than traditional approaches.
However, as these architectures make fewer structural assumptions, they typically require larger training datasets to learn accurate models, while they also lack robustness and are very sensitive to noise and perturbations.
A common problem in time series forecasting with deep neural networks is the one where the model just replicates the last observed value of the time series.
This is quite common in the case of noisy datasets, and is a problem of paramount importance since most real-world datasets contain noise.
In fact, this problem which we refer to as ``mimicking'' also depends on the nature of the employed loss function (\eg  MSE).

In this paper, our goal is to formally define the aforementioned problem such that practitioners can identify whether their models replicate previous values instead of making predictions. 
Therefore, we provide a definition of ``mimicking'' in time series forecasting and a methodology to quantify the extent to which a model suffers from it.  
Furthermore, we present examples of forecasts where this phenomenon is clearly observed.
The key technical contribution of this work is a carefully designed regularization term which can be added to the loss function and mitigate the drawbacks of ``mimicking'' which might occur in models trained by minimizing common loss functions.
The proposed regularization term is evaluated on a range of different datasets.
Our results suggest that the proposed term mitigates ``mimicking'' and reduces its impact on the model's performance.
The main contributions of this paper are summarized as follows:\\

\begin{itemize}
	\item To the best of our knowledge, we are the first to formally define the problem of ``mimicking'' in time series forecasting.
	\item We designed a regularization term that, when added to the loss function, mitigates to some degree the effect of ``mimicking''. 
	This term is general for all neural network architectures and does not make any assumptions.
	\item We specifically investigate and deal with the phenomenon of ``mimicking'' on three standard deep neural networks (LSTM, TCN and Transformer) which are some of the most widely used and effective models in time series forecasting and sequence modeling.
	\item The proposed regularization term improves the movement predictive performance of the vanilla models on $6$ public time series benchmark datasets and one stock dataset.
	On average, it leads to absolute improvements of $3.33\%$ in accuracy (considered for the three models), while MSE increases slightly.
\end{itemize}

\section{Related Work}\label{sec:related_work}

\paragraph{Time series forecasting.}
Before the advent of deep learning, the Auto-Regressive Integrated Moving Average (ARIMA) model~\cite{box2015time} and exponential smoothing~\cite{holt2004forecasting} were among the most popular and widely used methods for time series forecasting. 
However, these approaches have some drawbacks (\eg ARIMA assumes stationarity, while most real-world time series are not stationary), and thus have been replaced recently with neural network architectures~\cite{lim2021time}.
Different instances of recurrent neural networks such as Long Short Term Memory Networks (LSTMs)~\cite{hochreiter1997long} and Gated Recurrent Units (GRUs)~\cite{cho2014learning} have become the dominant approaches for time series forecasting mainly due to their ability to model complex patterns and long term dependencies, and to extract useful features from raw data.
Besides recurrent neural networks, convolutional neural networks have also been recently investigated in the task of time series forecasting.
The Temporal Convolution Network (TCN)~\cite{bai2018empirical} is perhaps the most prominent example from this family of models.
Attention mechanisms have proven very successful in many tasks and have also been applied to the problem of time series forecasting~\cite{qin2017dual,li2019enhancing}.
Different neural network components such as recurrent, convolutional and attention layers have been combined with autoregressive components to make predictions~\cite{lai2018modeling}.
The potential of residual connections along with a very deep stack of fully-connected layers in the context of time series forecasting has also been explored recently~\cite{oreshkin2020n}.
Matrix factorization methods have achieved prominent results in the case of high-dimensional time series data~\cite{yu2016temporal,sen2019think}.
Some recent works have combined neural networks and state space models~\cite{rangapuram2018deep,wang2019deep}.
Probabilistic forecasting, for predicting the distribution of possible future outcomes, has also recently started to receive increasing attention~\cite{salinas2020deepar,chen2020probabilistic}.

\paragraph{Loss functions.}
Besides the traditional functions (MSE, MAE, etc.), other measures that capture different time series properties have been proposed.
However, in most cases, these evaluation metrics are not differentiable, thus they cannot be directly employed as loss functions. 
Examples of such measures include the dynamic time warping algorithm which captures the shape of the time series, and standard evaluation metrics of supervised learning algorithms (\eg accuracy, f1-score) in the context of change point detection algorithms~\cite{aminikhanghahi2017survey}.
The need for measures alternatives to MSE has recently led to the development of new differentiable loss functions which capture different meaningful statistical properties of time series such as shape and time, including differentiable variants of dynamic time warping~\cite{cuturi2017soft,blondel2021differentiable}.
These differentiable dynamic time warping terms can also be combined with terms that penalize temporal distortions for more accurate temporal localization~\cite{guen2019shape}, while they have also been generalized to non-stationary time series~\cite{guen2020probabilistic} and binary series~\cite{rivest2019new}.

\section{The Phenomenon of ``Mimicking'' and How to Mitigate}\label{sec:methodology}

We first introduce some key notations for time series forecasting.
Let $x_{1:\tau} = (x_1,x_2,\ldots,x_\tau)$ be a univariate time series where $x_t \in \mathbb{R}$ denotes the value of the time series at time $t$.
The goal of a forecasting model is to predict the future values of the time series $z_{1:n} = (z_1, z_2, \ldots, z_n) = (x_{\tau+1}, x_{\tau+2}, \ldots, x_{\tau+n})$.
Let $\hat{z}_{1:n} = (\hat{z}_1, \hat{z}_2, \ldots, \hat{z}_n)$ denote the predictions of the forecasting model.
Neural network models for time series forecasting are typically trained to minimize the MSE which is defined as the sum of squared distances between the target variable and predicted values, \ie $\text{MSE} = \nicefrac{1}{n}\sum_{i=1}^{n}(\hat{z}_i-z_i)^2$.
Similar metrics, that measure the difference between the forecast and the actual value per time-step, such as MAE, are also employed in various applications.

\subsection{``Mimicking'' in Time Series Forecasting}

Even though MSE and related functions enjoy some nice properties (\eg MSE is convex on its input), when dealing with real-world data with multiple co-occurring patterns and noisy components, these functions might become sensitive to noise.
This might result into the problem of predicting previously seen values (usually the last seen observation in the time series), rather than making predictions based on long-term extracted patterns.
We next formalize the problem described above.
The following analysis focuses on the MSE loss, but it also applies to other loss functions that are commonly employed in time series forecasting (\eg MAE).
To investigate whether the model just replicates the last observed value of the time series, we can examine if the MSE between the forecast in time-step $t$ and the real value in time-step $t$ is greater than the MSE between the forecast in time-step $t$ and the real value in time-step $t-1$. 

\begin{definition}[``Mimicking'' in Time Series]
    We say that the phenomenon of ``mimicking'' in time series forecasting occurs if the following inequality holds
    \begin{equation}
        \sum_{i=1}^{n}(z_i - \hat{z}_i)^2 > \sum_{i=1}^{n} (z_{i-1} - \hat{z}_i)^2
        \label{eq:ineq}
    \end{equation}
    We can quantify the amount of ``mimicking'' as follows (the larger the (positive) value of $\text{MIM}$, the larger its severity): $\text{MIM} = \sum_{i=1}^{n} \big[(z_i - \hat{z}_i)^2 - (z_{i-1} - \hat{z}_i)^2 \big]$. 
\end{definition} 

\begin{figure}[t]
\begin{minipage}{0.45\linewidth}
    \centering
    \scriptsize
    \captionof{table}{MSE and ``mimicking'' as a function of the level of noise added to a synthetic dataset.}
    \begin{tabular}{lccc}
     & $\sigma$ & MSE $(\times 10^{-3})$ & MIM $(\times 10^{-3})$\\
     \hline
     \parbox[t]{2mm}{\multirow{5}{*}{\rotatebox[origin=c]{90}{LSTM}}} & 0 & 6.191 & 0.048 \\
     & 0.01 & 3.225 & 0.035 \\
     & 0.1 & 9.765 & 0.052 \\
     & 0.25 & 10.446 & 0.066 \\
     & 0.5 & 13.357 & 0.069 \\ \hline
     \parbox[t]{2mm}{\multirow{5}{*}{\rotatebox[origin=c]{90}{TCN}}} & 0 & 0.024 & -0.006\\
     & 0.01 & 0.007 & -0.003 \\
     & 0.1 &0.046 & 0.002 \\
     & 0.25 &0.068 & 0.004 \\
     & 0.5 &0.182 & 0.008 \\
    \end{tabular}
    \label{tab:synthetic}
\end{minipage}
\begin{minipage}{0.45\linewidth}
    \centering
    \includegraphics[scale=0.35]{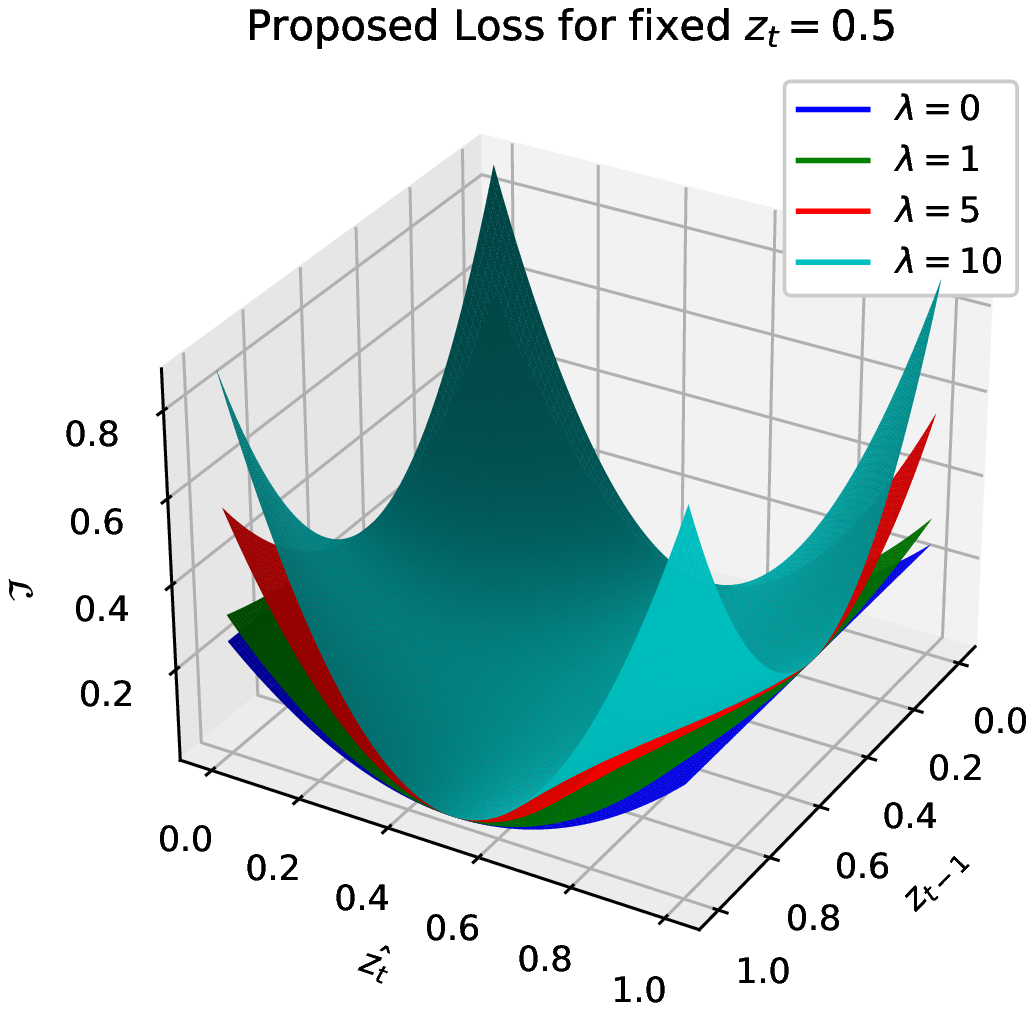}\\
    \captionof{figure}{A visualization of the proposed loss $\mathcal{L}$ of Equation~\eqref{eq:constraint_8} for different values of $\lambda$, $\hat{z}_t$ and $z_{t-1}$.}
    \label{fig:loss_func}
\end{minipage}
\end{figure}

To demonstrate that ``mimicking'' is related to the level of noise present in a dataset, we generated a synthetic dataset that corresponds to a sum of sinusoidal series with added random Gaussian noise (more details are given in section~\ref{sec:experimental_evaluation}).
A linear term is also added to the above terms.
Table~\ref{tab:synthetic} illustrates the MSE achieved by an LSTM and a TCN model along with the amount of ``mimicking'' as a function of the level of noise (\ie increasing variance).
We observe that the LSTM model is more prone to ``mimicking'' than the TCN model, while the greater the value of the variance of the Gaussian noise, the greater the impact of ``mimicking'' on the models' performance.
We also need to mention that the TCN model does not suffer from ``mimicking'' for $\sigma=0$ and $\sigma=0.01$.

\subsection{Proposed Regularization Term}

To mitigate the effects of mimicking in time series forecasting, we begin our analysis from the definition provided above.
Specifically, we would like the second term of inequality~\eqref{eq:ineq} to be greater or at least equal to the first term, \ie we would like the following to hold
\begin{equation}
    \sum_{i=1}^{n} \big[(z_i - \hat{z}_i)^2 - (z_{i-1} - \hat{z}_i)^2 \big] \leq 0
    \label{eq:constraint_1}
\end{equation}
By introducing the above term into the loss function, we directly punish ``mimicking'' to some extent.
However, incorporating solely the above term into the loss function gives rise to an unbounded function.
Indeed, in case $\sum_{i=1}^{n} (z_i - z_{i-1}) < 0$, setting $\hat{z}_i \rightarrow +\infty$ can drive the loss to negative infinity.
Likewise, if $\sum_{i=1}^{n} (z_i - z_{i-1}) > 0$, setting $\hat{z}_i \rightarrow -\infty$ also leads to a loss function that is unbounded from below.
Hence, since the loss is not bounded, there is no admissible estimator, and this will render the model to be of no practical use.

Note that a perfect model would achieve an MSE equal to $0$, \ie $\sum_{i=1}^n (z_i - \hat{z}_i) = 0$.
In such a scenario, we would like the loss function to take its lowest value.
If we replace the term that corresponds to the MSE in Equation~\eqref{eq:constraint_1} with $0$, we obtain
$0 - \sum_{i=1}^n (z_{i-1} - \hat{z}_i)^2 = -\sum_{i=1}^n (z_{i-1} - z_i)^2$
The equality is due to the fact that the model is perfect, \ie $z_i=\hat{z}_i$ $\forall i \in {1,\ldots, n}$ holds.
We would like the above term to be the lower bound of the proposed loss function (since the model is perfect).
Therefore, we have
\begin{equation}
    -\sum_{i=1}^n (z_{i-1} - z_i)^2 \leq \sum_{i=1}^n(z_i - \hat{z}_i)^2 - \sum_{i=1}^n (z_{i-1} - \hat{z}_i)^2
    \label{eq:constraint_2}
\end{equation}
By combining Equations~\eqref{eq:constraint_1} and~\eqref{eq:constraint_2}, we obtain the following inequality
\begin{equation}
    \begin{split}
        -\sum_{i=1}^n (z_{i-1} - z_i)^2 \leq \sum_{i=1}^n(z_i - \hat{z}_i)^2 &- \sum_{i=1}^n (z_{i-1} - \hat{z}_i)^2 \leq 0 \\
        \iff 0 \leq \sum_{i=1}^n 2(z_i - z_{i-1})(z_i - \hat{z}_i) &\leq \sum_{i=1}^n (z_{i-1} - z_i)^2 \\
        \iff 0 \leq \sum_{i=1}^n (z_i - z_{i-1})(z_i - \hat{z}_i) &\leq \frac{1}{2} \sum_{i=1}^n (z_{i-1} - z_i)^2 \\
    \end{split}
\end{equation}
Ideally, we would like the above inequality to hold.
That would mean that the phenomenon of mimicking does not occur.
However, the middle term is still not bounded, thus we cannot directly minimize that term. 
Note that all the terms are nonnegative.
Hence, we can square all the sides of the inequality as follows
\begin{equation}
    0 \leq \sum_{i=1}^n \big[(z_i - z_{i-1})(z_i - \hat{z}_i)\big]^2 \leq \frac{1}{4} \sum_{i=1}^n (z_{i-1} - z_i)^4
\end{equation}
Now, the middle term is nonnegative by construction, and we can thus safely minimize it.
Interestingly, the above function is continuous and differentiable which are both desirable properties for loss functions.
For instance, the first and second derivatives of the function are shown below

\begin{equation}
    \frac{d}{d\hat{z}_i} \sim \sum_{i=1}^n 2(z_{i-1}-z_i)^2(\hat{z}_i-z_i), \qquad \frac{d^2}{d\hat{z}_i^2} \sim \sum_{i=1}^n 2(z_{i-1}-z_i)^2
\end{equation}

From the above, it is also clear that the second derivative of the function is nonnegative on its entire domain, thus the function is convex.
However, we need to mention that even though the function is convex in $\hat{z}_i$, in case neural networks are employed (or other non-linear models), we have $\hat{z}_i = f(z_{i-1}, \ldots, z_{i-k} ; \theta)$ and the function is not convex in $\theta$.

Our proposed loss function for a sequence of $n$ time-steps is defined as 
\begin{equation}
    \mathcal{L} = \sum_{i=1}^n (z_i-\hat{z}_i)^2 + \lambda \sum_{i=1}^n \big[(z_i - z_{i-1})(z_i - \hat{z}_i)\big]^2
    \label{eq:constraint_8}
\end{equation}
where $\lambda$ is a parameter which controls the importance of the regularization term, \ie how much penalty needs to be imposed to alleviate ``mimicking''.
The two factors $(z_i - z_{i-1})$, $(z_i - \hat{z}_i)$ that constitute the penalty term above can be interpreted as a discrete-time cross-correlation measure function between the difference of the series at $i$ and $i-1$ and the predicted error at $i$. 
If we expand the term for a specific $i$, we derive $[(z_i - z_{i-1})z_i - (z_i - z_{i-1})\hat{z}_i)]^2$. 
The closer the prediction $\hat{z}_i$ is to $z_{i-1}$ and the farther $z_i$ is from $z_{i-1}$, the larger the imposed penalty term will be.
Figure~\ref{fig:loss_func} illustrates how the proposed loss function varies as a function of $z_{t-1}$ and $\hat{z}_t$ for different values of $\lambda$ (for $z_t$ fixed to $0.5$).

In some cases, the model might not replicate solely the last observed value of the time series $x_\tau$, but also observations that occurred farther in the past, \eg $x_{\tau-1}, x_{\tau-2}$, etc.
We next generalize the proposed penalty term to account for such kind of scenarios.
To prevent a neural network model from replicating the last $K$ observations, we can use the following loss function
\begin{equation}
    \mathcal{L} = \sum_{i=1}^n (z_i-\hat{z}_i)^2 + \lambda \sum_{i=1}^n \sum_{k=1}^K \big[(z_i - z_{i-k})(z_i - \hat{z}_i)\big]^2
    \label{eq:constraint_9}
\end{equation}

The proposed loss of Equation~\eqref{eq:constraint_8} can also be generalized to the case of multi-step ahead forecasting.
Specifically, it can be directly applied to iterative $1$-step methods~\cite{lim2021time}, while in the case of direct multi-horizon forecasting (vector output/Seq2Seq architectures), we need to consider vectors $(\hat{z}_1, \hat{z}_2, \ldots, \hat{z}_h)$ of length $h$ which refer to the desired horizon.
Direct multi-horizon strategies have been recently preferred, despite the flexibility of iterative ones.

\section{Experimental Evaluation}\label{sec:experimental_evaluation}

\subsection{Datasets}
\noindent\textbf{Synthetic.}
This synthetic dataset corresponds to a sum of sinusoidal series with added random noise:
$y(t) = \sin{(t)} + \sin{\Big(\frac{\pi}{2}t\Big)} + \sin{\Big(\frac{-3\pi}{2}t\Big)} + \epsilon(t)$,  where $\epsilon(t)$ is a Gaussian distribution with mean $\mu$ and variance $\sigma^2$ ($\mu=0$, $\sigma=0.5$).

\noindent\textbf{Monthly sunspots.}
This dataset describes a monthly count of the number of observed sunspots from $1749$ to $1983$, a total of $2,820$ observations.

\noindent\textbf{Electricity.}
It contains electricity consumption measurements (kWh) from $321$ clients, recorded every $15$ minutes from $2012$ to $2014$.
We utilize the first univariate series of length $26,304$.

\noindent\textbf{Beijing PM2.5.}
This hourly dataset contains the PM2.5 data of the US Embassy in Beijing.
It is a multivariate dataset that consists of eight variables, including the PM2.5 concentration and a total number of $43,824$ observations.
The task is to predict the future hourly concentration given the other variables.

\noindent\textbf{Solar Energy.}
It contains the solar power production data from photovoltaic plants in Alabama in 2006. We utilize the first univariate series of length $26,304$.

\noindent\textbf{Exchange Rate.} It includes the exchange rates of eight foreign countries (Australia, Britain, Canada, China, Japan, New Zealand, Singapore, and Switzerland) from 1990 to 2016.  We utilize the first univariate series of length $7,588$.

\subsection{Evaluation Metrics}

In order to evaluate the performance of our proposed loss function in the experiments that follow, we employ the following metrics:\\
\noindent\textbf{Mean Squared Error (MSE) and Shifted Mean Squared Error (s-MSE).}
MSE compares the predictions $\hat{z_t}$ against the targets $z_t$.
Shifted MSE compares $\hat{z_t}$ against the last values $z_{t-1}$, \ie $\text{s-MSE} = \nicefrac{1}{n} \sum_{i=1}^{n}(\hat{z}_i-z_{i-1})^2$.

\noindent\textbf{Accuracy (Acc) and Shifted Accuracy (s-Acc).}
To compute these two metrics, we turn the forecasting problem into a classification one.
Let $v$ be a $n$-dimensional vector such that its $i$-th element is defined as $v_i = \text{change}(z_i, z_{i-1})$ where $\text{change}(a, b) = \text{sign}(a - b)$.
Let also $\hat{v}$ be a $n$-dimensional vector such that $\hat{v}_i = \text{change}(\hat{z}_i, \hat{z}_{i-1})$.
Then, Acc is defined as the accuracy between the above two vectors.
s-Acc is defined as the accuracy between the vector $\hat{v}$ and the vector $v$ shifted by $1$ step to the left.
Predicting whether the value of a time series will increase or decrease is a task of high importance for many applications such as stock price prediction.
Indeed, successful predictions would enable hedge funds or investors lay a successful strategy for buying and selling stocks.

We should note here that MSE and accuracy are two metrics orthogonal to each other.
A time series forecasting model ideally would achieve a low value of MSE and a high accuracy.
Models that suffer from ``mimicking'' can yield low values of MSE, thus achieving solely a low MSE might not be a clear indicator of the model's predictive power.
On the other hand, a model that yields solely high accuracy captures the shape and the change points of the time series, but the predicted values might significantly deviate from the actual values of the series.

\begin{figure}[t]
\begin{minipage}{0.25\textwidth}
\centering
\subfloat[Synthetic-LSTM]{\label{examples_mimicking:a}\includegraphics[width=\linewidth]{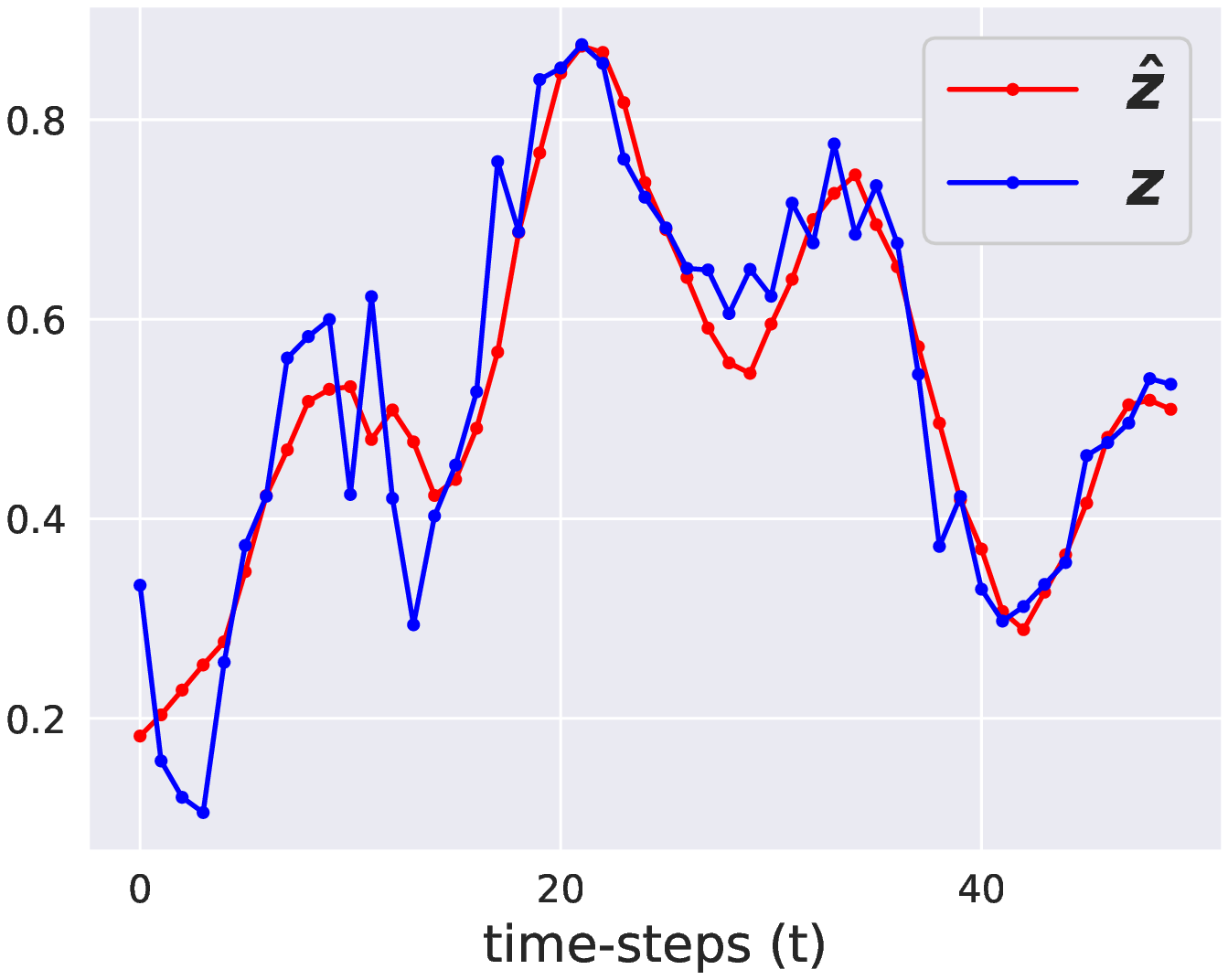}}
\end{minipage}%
\begin{minipage}{.25\textwidth}
\centering
\subfloat[Sunspots-LSTM]{\label{examples_mimicking:b}\includegraphics[width=\linewidth]{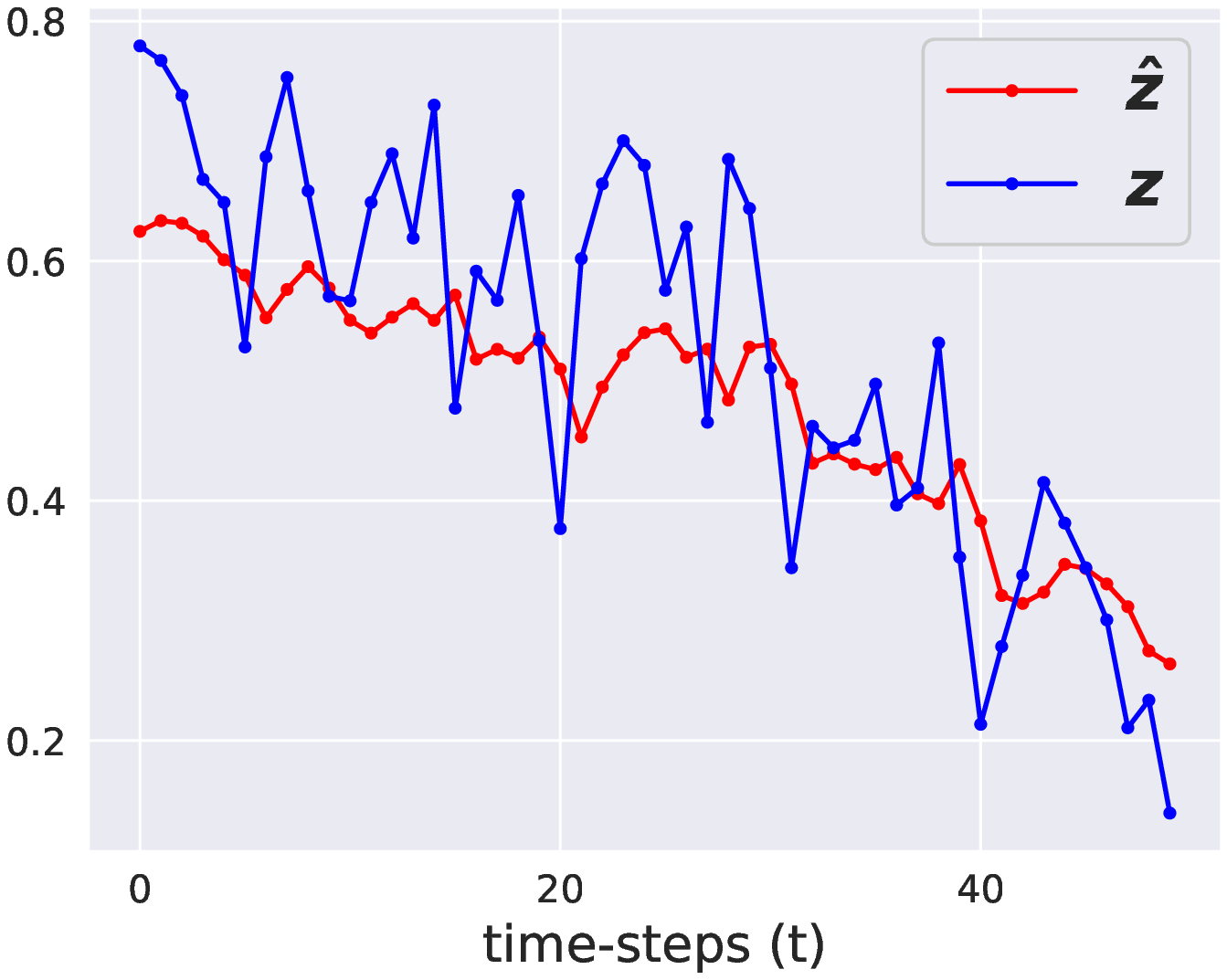}}
\end{minipage}%
\begin{minipage}{.25\textwidth}
\centering
\subfloat[Electricity-TCN]{\label{examples_mimicking:c}\includegraphics[width=\linewidth]{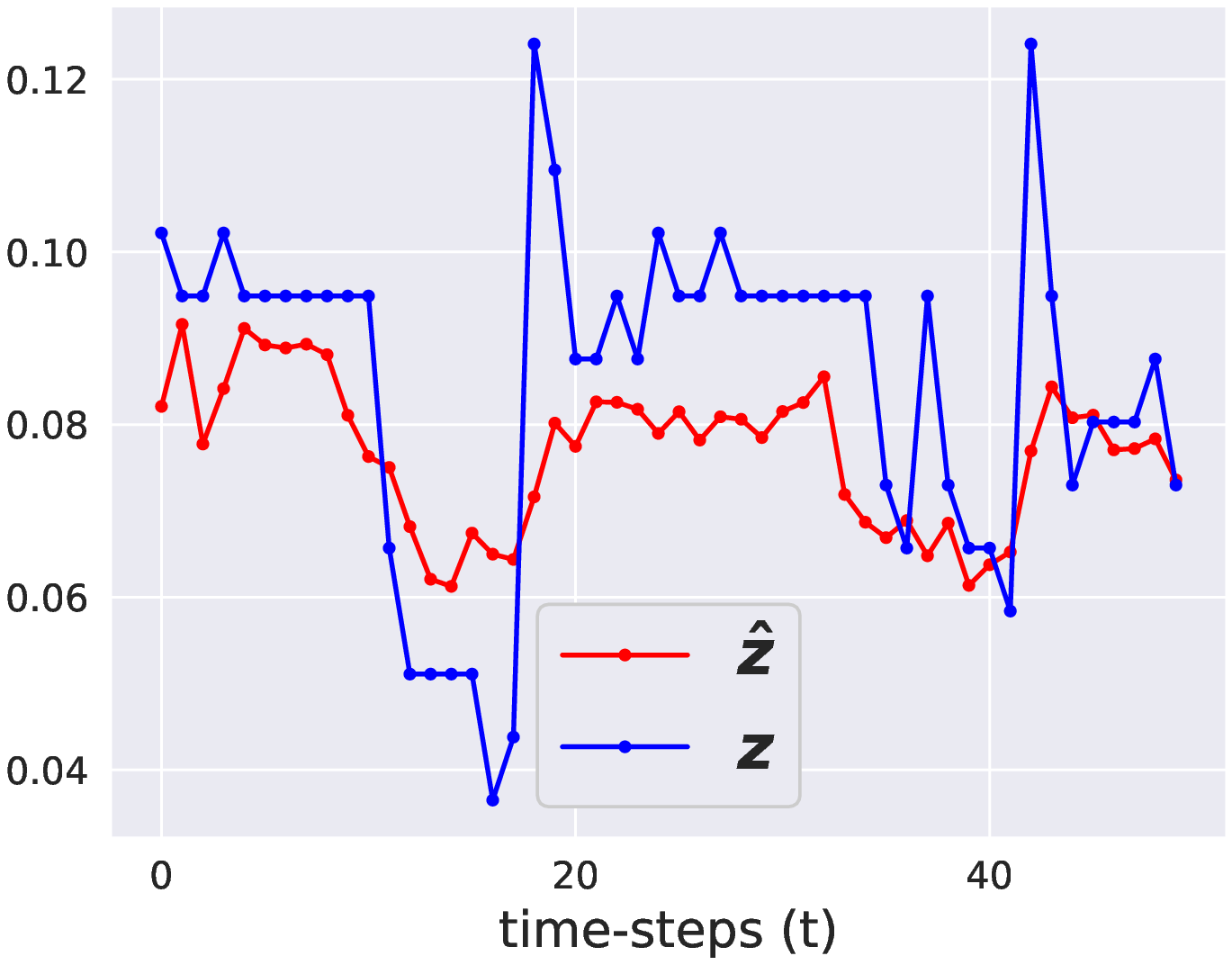}}
\end{minipage}%
\begin{minipage}{.25\textwidth}
\centering
\subfloat[PM2.5-TCN]{\label{examples_mimicking:d}\includegraphics[width=\linewidth]{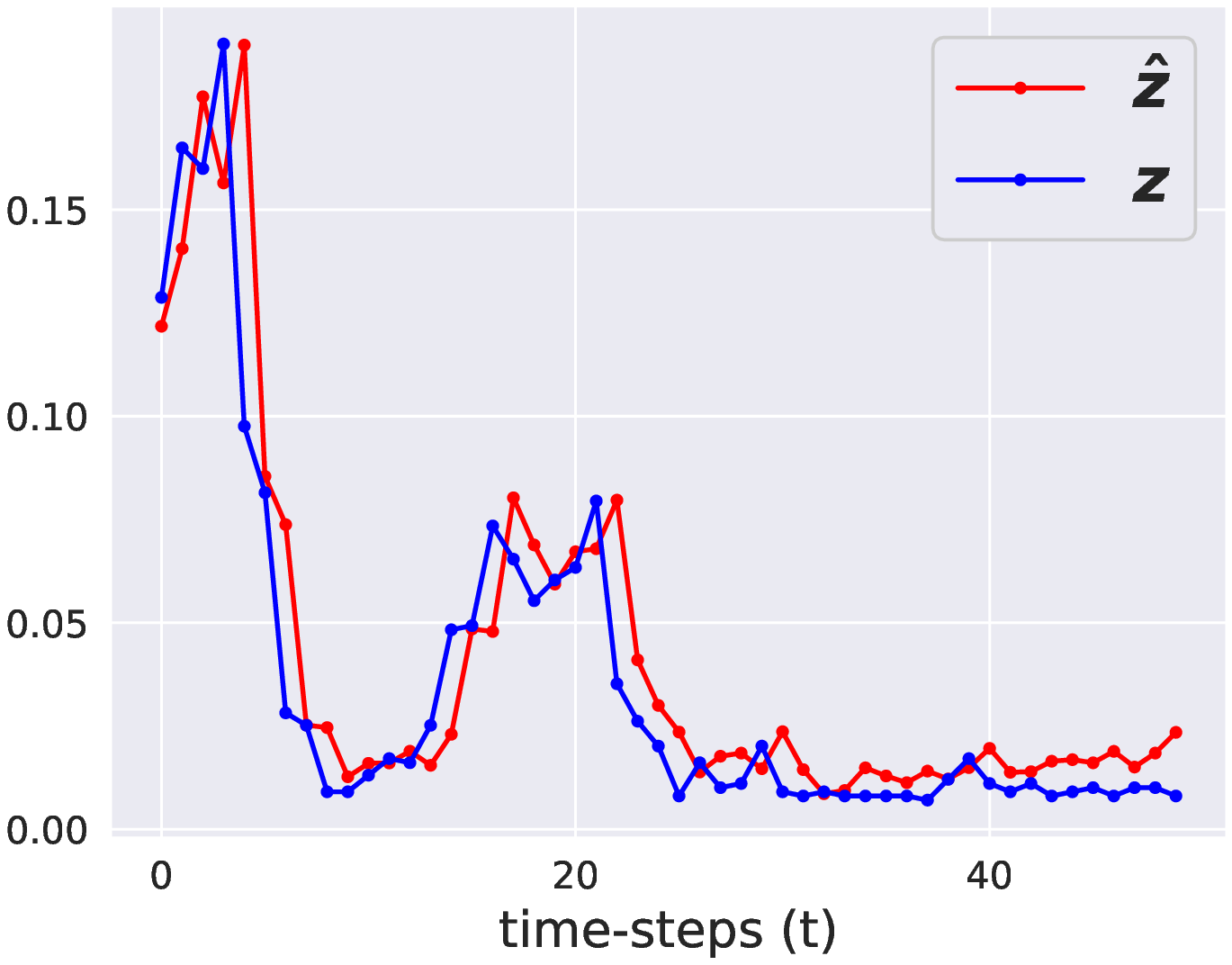}}
\end{minipage}\par\medskip
\caption{$1$-step predictions of models trained with MSE on part of the test set.}
\label{fig:examples_mimicking}
\end{figure}

\begin{table}[t]
    \centering
    \tiny
    \caption{$1$-step prediction performance of the different models, and the baseline on the $6$ considered datasets. All MSE/s-MSE results are in scale $(\times 10^{-3})$.}
    \def\arraystretch{1.6}
    \begin{tabular}{|c|c||cc|cc|cc|cc|cc|cc|} \hline
    \multicolumn{2}{|l||}{\multirow{2}{*}{\bf{Methods}}} & \multicolumn{2}{c|}{\textbf{Synthetic}} & \multicolumn{2}{c|}{\textbf{Sunspots}} & \multicolumn{2}{c|}{\textbf{Electricity}} & \multicolumn{2}{c|}{\textbf{Beijing PM2.5}} &
    \multicolumn{2}{c|}{\textbf{Solar}} & \multicolumn{2}{c|}{\textbf{Exchange Rate}}\\
    \cline{3-14}
    \multicolumn{2}{|l||}{} & \textbf{MSE} & \textbf{s-MSE} & \textbf{MSE} & \textbf{s-MSE} & \textbf{MSE} & \textbf{s-MSE} & \textbf{MSE} & \textbf{s-MSE} & \textbf{MSE} & \textbf{s-MSE} & \textbf{MSE} & \textbf{s-MSE}\\
    \hline
    \multirow{5}{*}{Avg. Window} & $1$ & 9.759 & 0.0 & 6.825 & 0.0 & 5.063 & 0.0 & 0.489 & 0.0 & 1.803 & 0.0 & 0.248 & 0.0 \\
     & $3$ & 12.398 & 4.524 & 6.881 & 3.03 & 4.766 & 2.15 & 0.905 & 0.30 & 3.531 & 1.150 & 0.244 & 0.103\\
     & $5$ & 18.378 & 9.791 & 7.541 & 4.54 & 4.851 & 3.05 & 1.358 & 0.72 & 5.654 & 2.918 & 0.292 & 0.171\\
     & $7$ & 22.896 & 15.382 & 8.642 & 5.91 & 5.066 & 3.62 & 1.793 & 1.16 & 7.998 & 4.998 & 0.348 & 0.235\\
     & $9$ & 24.883 & 19.127 & 9.737 & 7.22 & 5.363 & 4.11 & 2.192 & 1.57 & 10.485 & 7.291 & 0.407 & 0.298\\ \hline
    \multicolumn{2}{|c||}{LSTM} & 4.742 & 4.778 & 5.739 & 3.572 & 3.594 & 1.49 & 0.421 & 0.08 & 1.358 & 0.463 & 0.388 & 0.236 \\ \hline
    \multicolumn{2}{|c||}{TCN} & 3.784 & 5.650 & 6.043 & 3.485 & 3.582 & 2.22 & 0.506 & 0.12 & 1.489 & 0.397 & 0.270 & 0.122 \\ \hline
    \multicolumn{2}{|c||}{Transf.} & 5.422 & 5.128 & 12.21 & 11.21 & 4.299 & 2.659 & 0.560 & 0.18 & 1.802 & 0.391 & 3.811 & 3.747 \\
    \hline
    \end{tabular}
    \label{tab:examples_mimicking}
\end{table}

\subsection{Experimental Setup}

We divide the sequence into multiple samples, where $T$ observations are given as input and the expected output is the actual value of $n$ observations that follow each one of those $T$ observations.
We choose $T$ from $\{ 32, 64, 128, 256, 512 \}$.

We choose parameters as follows.
For the LSTM model, we use a single LSTM layer.
We use the hidden state of the last time step of the LSTM layer as the vector representation of the time series. The generated vector representations are then fed into a two layer MLP with a ReLU activation function.
For the TCN model, we adjust the parameters to capture the different history lengths $T$ that we test, from the equation $R_{field}=2^{D-1}\cdot K_{size}$, for $K_{size}=2$ and $D$ the number of dilation layers.
Each layer has dilation rate of $2^{N_l-1}$, where $N_l=\{1,2,...,D\}$.  
We also implement a model consisting of two stacked encoders of the Transformer architecture followed by a fully-connected layer for the final prediction. 
The hidden-dimension size of the LSTM, TCN and Transformer layers is chosen from $\{32, 64, 128, 256\}$.
For all the three models, we use the Adam optimizer with an initial learning rate of $10^{-3}$ and decay the learning rate by $0.1$ every $10$ epochs.
We choose the batch size from $\{32, 64, 128\}$.
We set the number of epochs to $100$, and we retrieve the model that achieves the lowest validation loss.
The regularization parameter $\lambda$ is chosen from $\{0.1, 0.5, 1, 5, 10, 20, 50, 100, 200, 500, 800, 1000\}$.

We also implement a simple baseline method (Avg. Window) which, given the past $n$ values of the time series, predicts the average value: $\hat{z}_{t} = \nicefrac{1}{n} \sum_{i=1}^n z_{t-i}$.

\subsection{Results}

\begin{table}[t]
    \centering
    \scriptsize
    \caption{$1$-step prediction performance of the different models on the $6$ considered datasets trained with MSE and with the proposed loss function. All MSE/s-MSE results are in scale $(\times 10^{-3})$. We mention in bold the maximum accuracy (Acc) and we underline the minimum shifted accuracy (s-Acc) achieved for each model.}
    \def\arraystretch{1.2}
    \begin{tabular}{|l||cccc|cccc|cccc|} \hline
    \multirow{2}{*}{\bf{Methods}} & \multicolumn{4}{c|}{\textbf{Synthetic}} & \multicolumn{4}{c|}{\textbf{Sunspots}} & \multicolumn{4}{c|}{\textbf{Electricity}} \\
    \cline{2-13}
    & \textbf{MSE} & \textbf{s-MSE} & \textbf{Acc} & \textbf{s-Acc} & \textbf{MSE} & \textbf{s-MSE} & \textbf{Acc} & \textbf{s-Acc} & \textbf{MSE} & \textbf{s-MSE} & \textbf{Acc} & \textbf{s-Acc} \\
    \hline
    LSTM & 4.742 & 4.778 & 0.616 & 0.695 & 5.739 & 3.572 & 0.388 & 0.739 & 3.594 & 1.490 & 0.278 & 0.702 \\ 
    LSTM+reg. & 5.747 & 9.767 & \textbf{0.657} & \underline{0.586} & 8.857  & 8.033 & \textbf{0.440} & \underline{0.628} & 28.97 & 29.79 & \textbf{0.440} & \underline{0.122} \\ \hline 
    TCN & 3.784 & 5.650 & 0.677 & 0.636 & 6.043 & 3.485 & 0.439 & 0.725 & 3.582 & 2.220 & 0.366 & 0.497 \\ 
    TCN+reg. & 4.491 & 9.335 & \textbf{0.685} & \underline{0.569} & 7.554 & 6.207 & \textbf{0.529} & \underline{0.631} & 26.01 & 27.32 & \textbf{0.442} & \underline{0.319} \\ \hline 
    Transf. & 5.422 & 5.128 & 0.603 & 0.713 & 12.21 & 11.21 & 0.447 & 0.659 & 4.299 & 2.659 & 0.350 & 0.509 \\ 
    Transf.+reg. & 6.467 & 8.780 & \textbf{0.636} & \underline{0.610} & 22.05  & 21.99 & \textbf{0.479} & \underline{0.488} & 15.86 & 16.73 & \textbf{0.417} & \underline{0.337} \\ \hline
    \end{tabular}\\
    \begin{tabular}{|l||cccc|cccc|cccc|} \hline
    \multirow{2}{*}{\bf{Methods}} & \multicolumn{4}{c|}{\textbf{Beijing PM2.5}} & \multicolumn{4}{c|}{\textbf{Solar}} & \multicolumn{4}{c|}{\textbf{Exchange Rate}} \\
    \cline{2-13}
    & \textbf{MSE} & \textbf{s-MSE} & \textbf{Acc} & \textbf{s-Acc} & \textbf{MSE} & \textbf{s-MSE} & \textbf{Acc} & \textbf{s-Acc} & \textbf{MSE} & \textbf{s-MSE} & \textbf{Acc} & \textbf{s-Acc} \\
    \hline
    LSTM  & 0.421 & 0.080 & 0.547 & 0.846 & 1.358 & 0.463 & \textbf{0.262} & 0.352 & 0.388 & 0.236 & 0.427 & 0.685 \\ 
    LSTM+reg. & 1.605  & 1.263 & \textbf{0.553} & \underline{0.661} & 3.638 & 3.250 & 0.257 & \underline{0.313} & 0.524  & 0.414 & \textbf{0.436} & \underline{0.594} \\ \hline
    TCN  & 0.506 & 0.120 & 0.540 & 0.822 & 1.489 & 0.397 & \textbf{0.269} & 0.364 & 0.270 & 0.122 & 0.430 & 0.643 \\ 
    TCN+reg. & 0.518 & 0.140 & \textbf{0.542} & \underline{0.809} & 3.855 & 3.240 & 0.255 & \underline{0.304} & 0.494 & 0.379 & \textbf{0.443} & \underline{0.586} \\ \hline
    Transf. & 0.560 & 0.180 & \textbf{0.559} & 0.791 & 1.802 & 0.391 & \textbf{0.266} & 0.363 & 3.811 & 3.747 & 0.454 & \underline{0.444} \\
    Transf.+reg. & 0.890  & 0.510 & 0.545 & \underline{0.696} & 3.927 & 2.834 & 0.254 & \underline{0.320} & 9.320  & 9.321 & \textbf{0.471} & 0.453 \\ \hline
    \end{tabular}\\
    \label{tab:regularization}
\end{table}

\begin{figure}[t]
\begin{minipage}{0.25\textwidth}
\centering
\subfloat[Synthetic-LSTM]{\label{regularized_mimicking:a}\includegraphics[width=\linewidth]{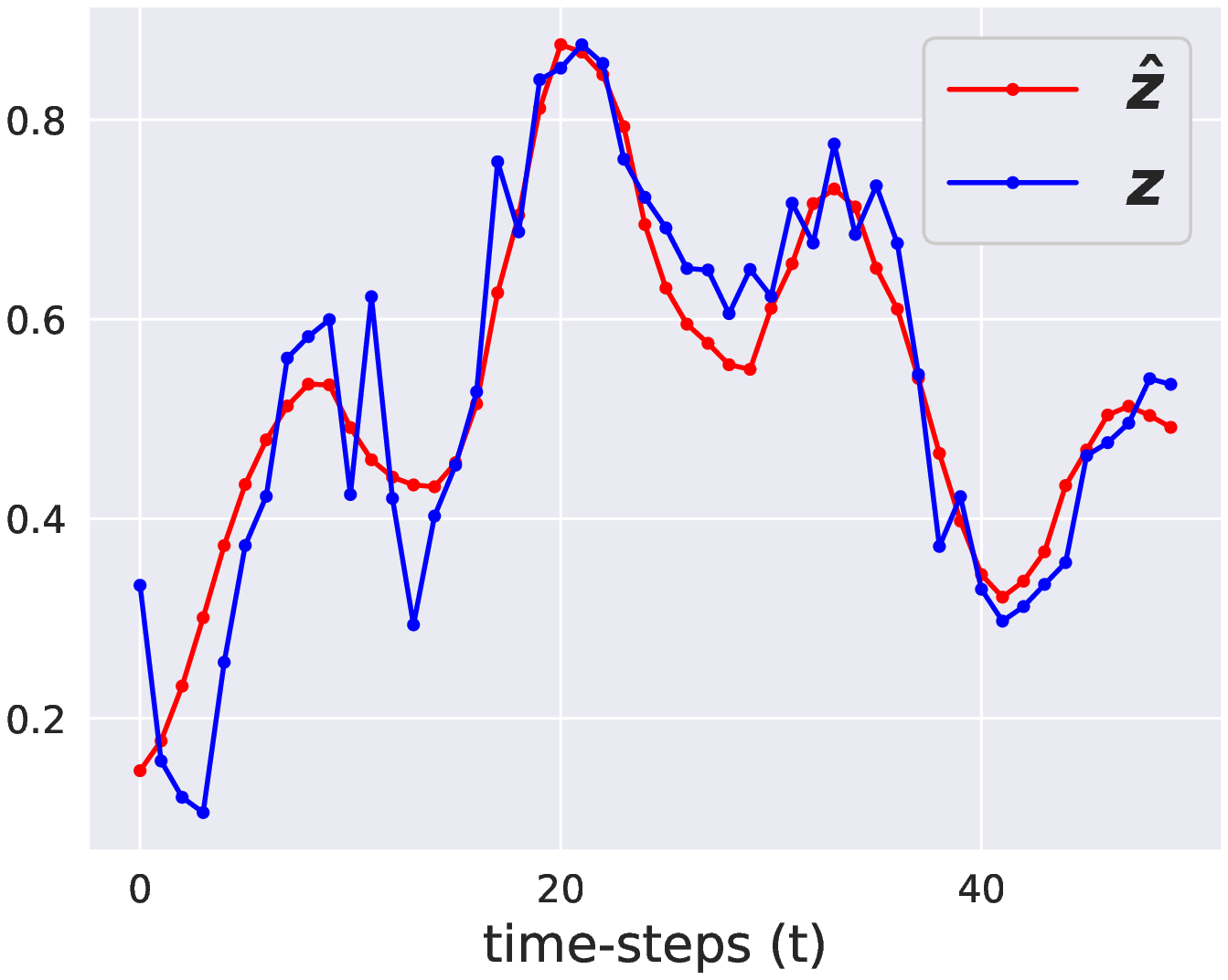}}
\end{minipage}%
\begin{minipage}{.25\textwidth}
\centering
\subfloat[Sunspots-LSTM]{\label{regularized_mimicking:b}\includegraphics[width=\linewidth]{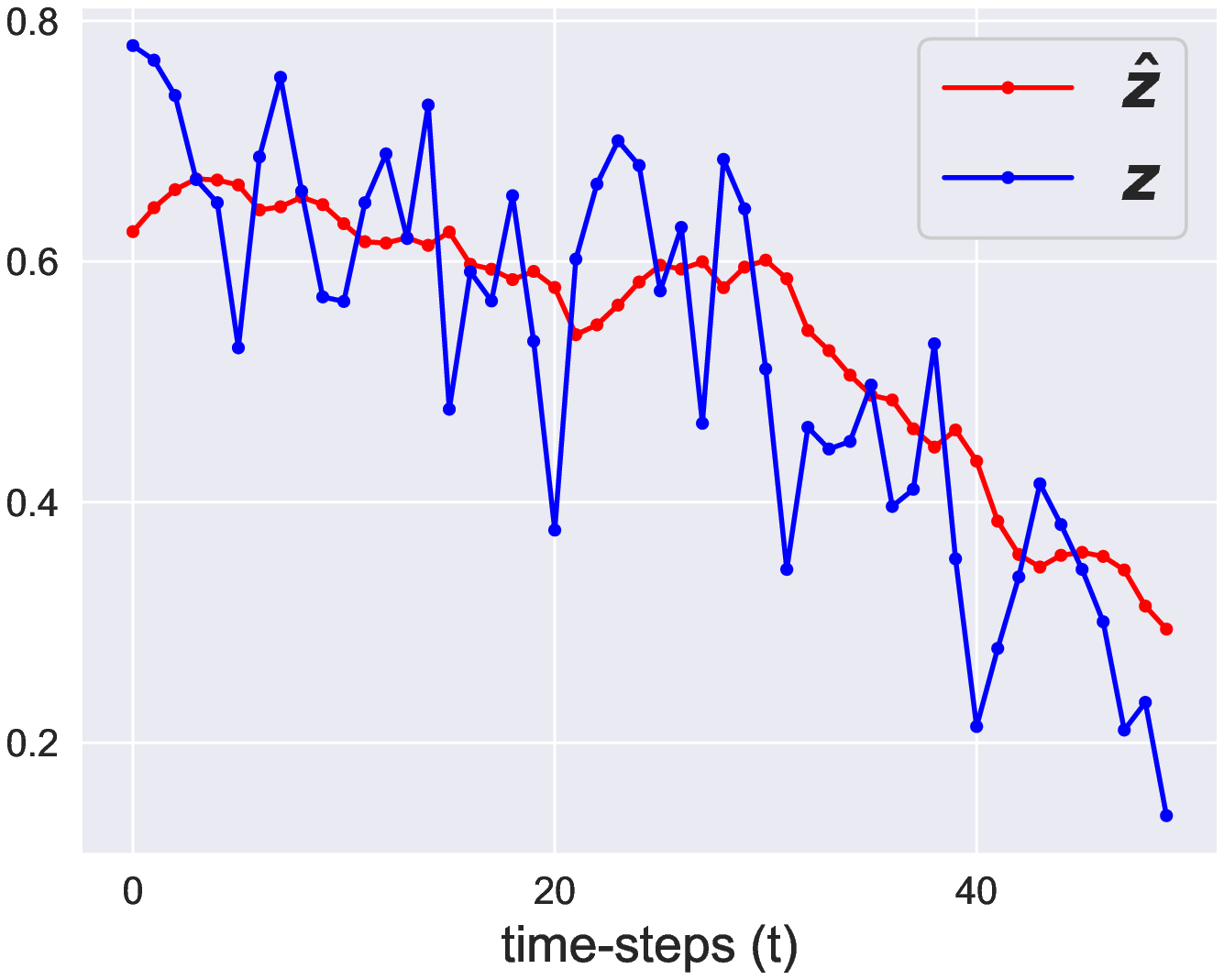}}
\end{minipage}%
\begin{minipage}{.25\textwidth}
\centering
\subfloat[Electricity-TCN]{\label{regularized_mimicking:c}\includegraphics[width=\linewidth]{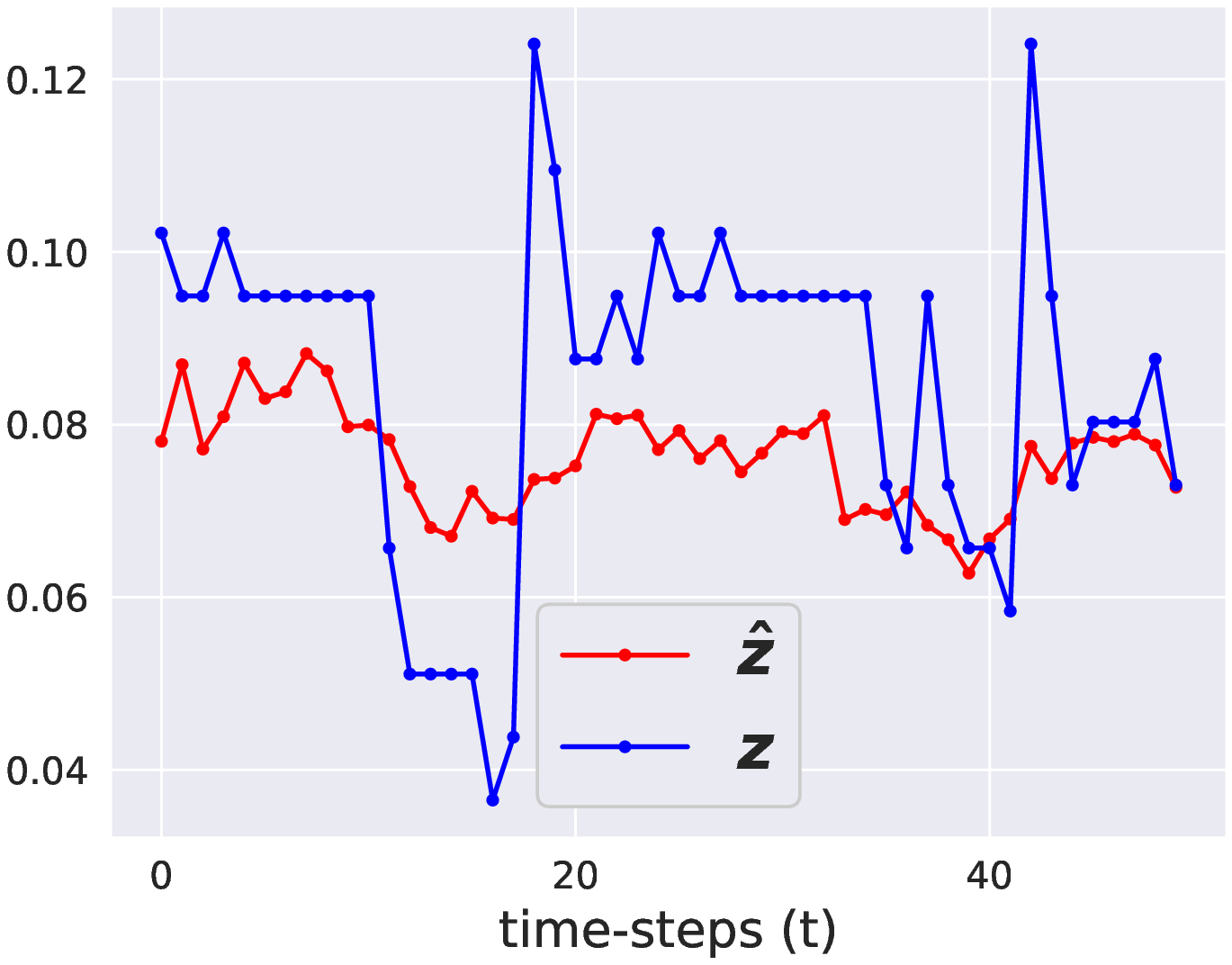}}
\end{minipage}%
\begin{minipage}{.25\textwidth}
\centering
\subfloat[PM2.5-TCN]{\label{regularized_mimicking:d}\includegraphics[width=\linewidth]{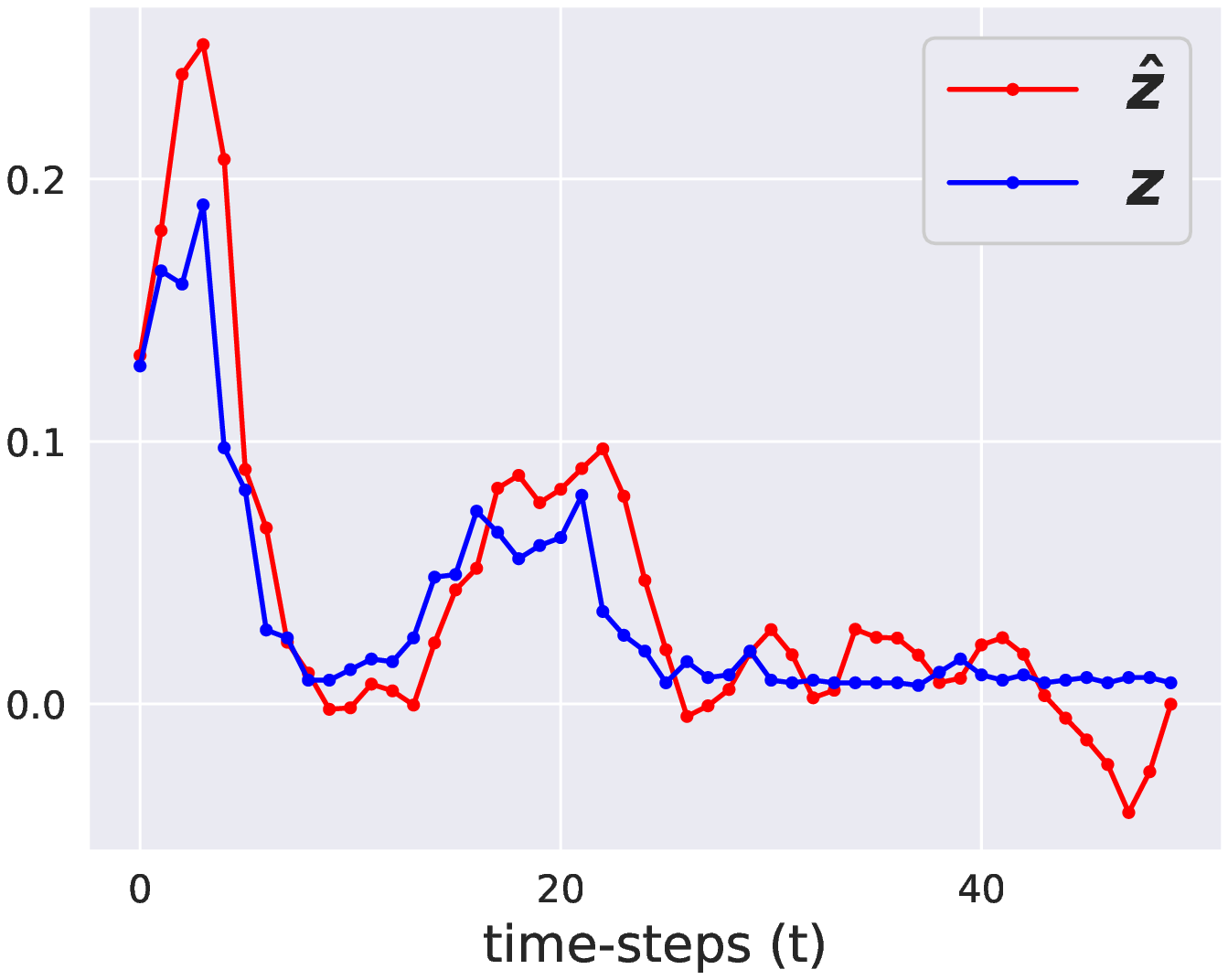}}
\end{minipage}
\caption{Predictions as in Figure~\ref{fig:examples_mimicking} but after training with the proposed loss.}
\label{fig:regularized_mimicking}
\end{figure}

\noindent\textbf{Examples of ``mimicking''.}
We next provide some examples of forecasts where the LSTM model and the TCN model just learn to replicate the last seen observations.
Figure~\ref{fig:examples_mimicking} illustrates such examples for some of the considered datasets.
The first $2$ plots (\ie (a) and (b)) correspond to predictions of the LSTM model, while the last $2$ plots (\ie (c) and (d)) to predictions of the TCN model.
On the synthetic dataset, the LSTM model learns to infer quite accurately the future values of the time series.
This is mainly due to the simplistic nature of that dataset.
On the other hand, on the real-world datasets, the two models fail to generalize, replicating previously observed data. 
This is especially true for plots (b) and (d).
Specifically, on the Beijing PM2.5 dataset, ``mimicking'' is observed to a very large extent, probably due to the complexity of the dataset. 

\begin{figure}[t]
\begin{minipage}{0.25\textwidth}
\centering
\subfloat[Synthetic-LSTM]{\label{lambdas:a}\includegraphics[width=\linewidth]{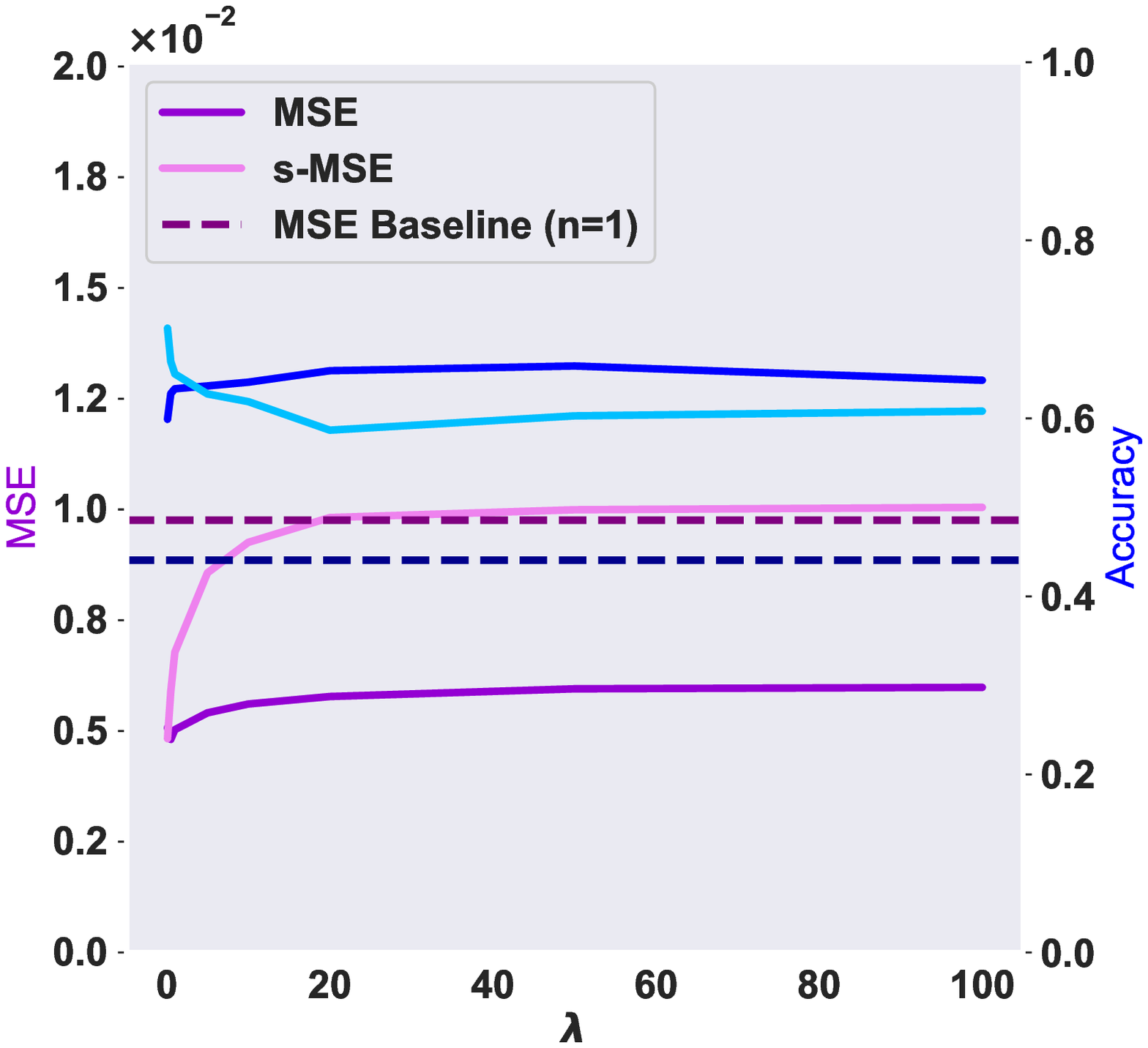}}
\end{minipage}%
\begin{minipage}{0.25\textwidth}
\centering
\subfloat[Electricity-LSTM]{\label{lambdas:b}\includegraphics[width=\linewidth]{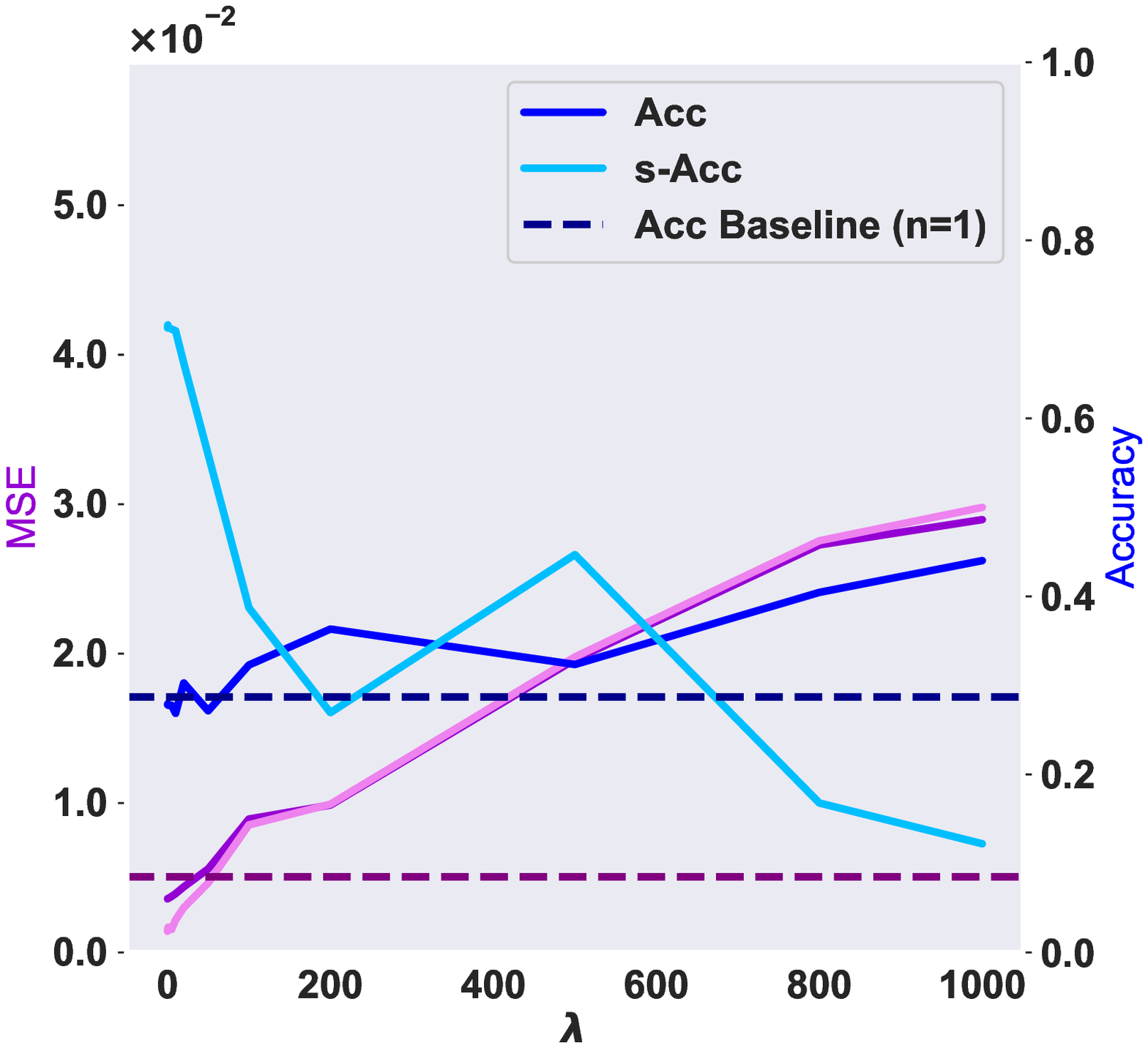}}
\end{minipage}%
\begin{minipage}{.25\textwidth}
\centering
\subfloat[PM2.5-LSTM]{\label{lambdas:c}\includegraphics[width=\linewidth]{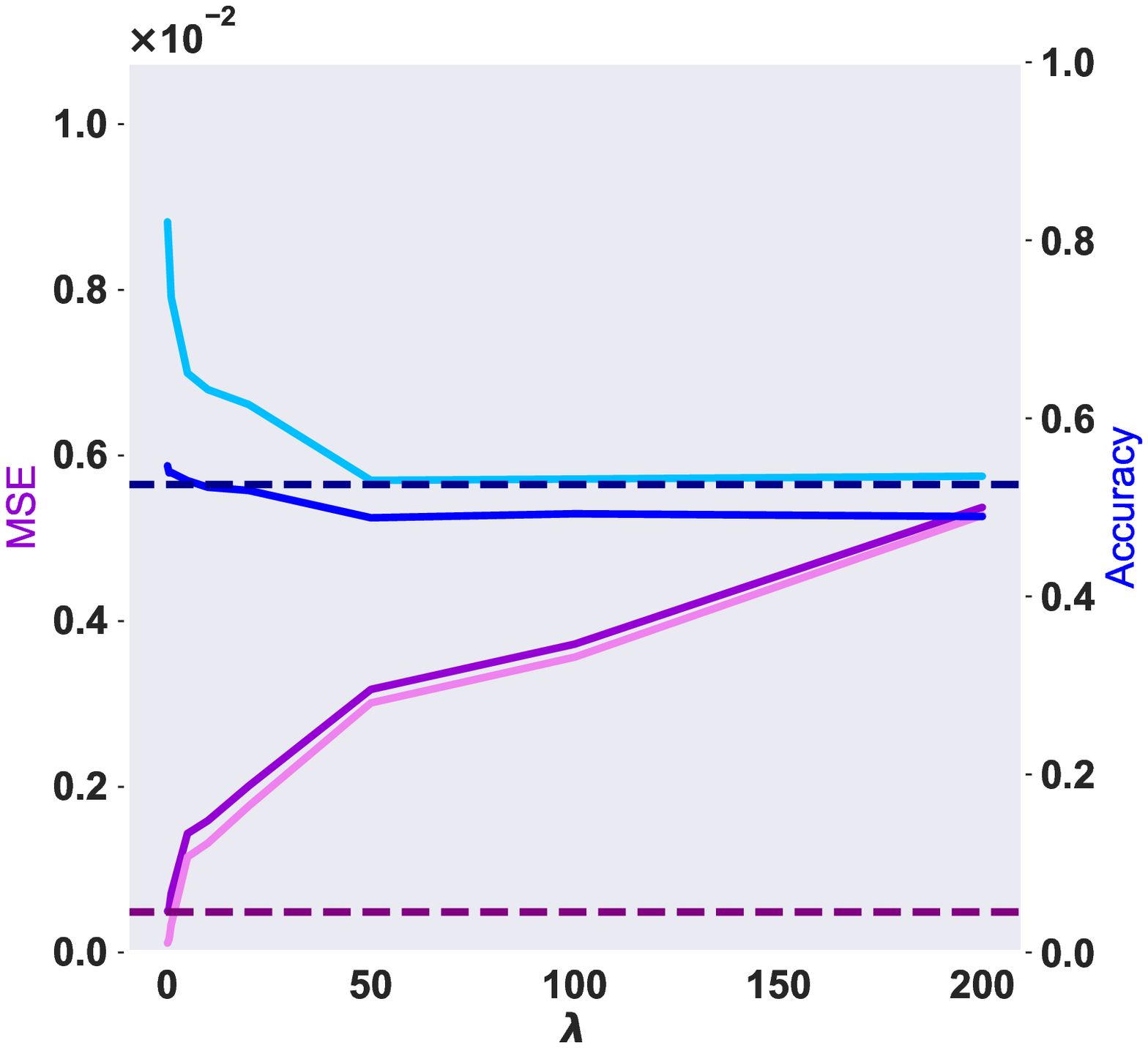}}
\end{minipage}%
\begin{minipage}{.25\textwidth}
\centering
\subfloat[Sunspots-TCN]{\label{lambdas:d}\includegraphics[width=\linewidth]{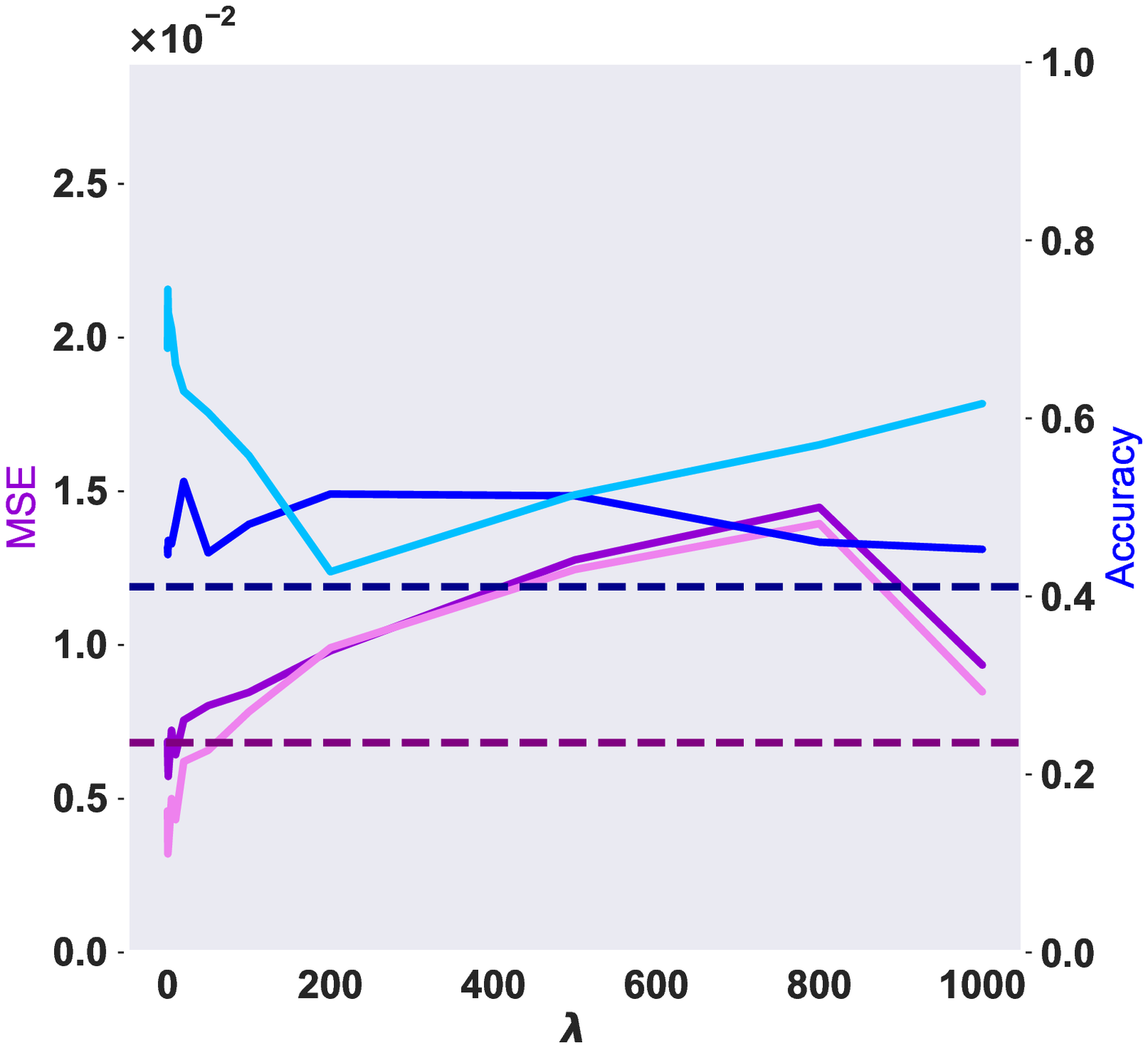}}
\end{minipage}
\caption{$1$-step prediction performance of different models, trained to minimize the proposed loss, as a function of $\lambda$ and the baseline (Avg. Window ($n=1$)).}
\label{fig:lambdas}
\end{figure}

Besides the above qualitative results, we also present some quantitative results in Table~\ref{tab:examples_mimicking}.
We can see that on the $5$ real-world datasets, the LSTM, TCN and Transformer models suffer from ``mimicking'' since s-MSE is smaller than MSE in all cases.
Interestingly, s-MSE can even be an order of magnitude smaller than MSE (see LSTM on Beijing PM2.5 and all three models on Solar).
On the other hand, on the synthetic dataset both the LSTM and the TCN model achieve a smaller MSE than s-MSE.
Thus, on this dataset, the two models are more robust. 
Indeed, this dataset is less noisy, while its trend is more predictable than that of the real-world datasets.
With regards to the baselines, in most cases, they also achieve low values of MSE (especially when $n=1$).
In fact, on the Beijing PM2.5 dataset, the Avg. Window ($n=1$) outperforms the TCN model since it yields a smaller MSE than TCN.
This interesting result indicates that a simplistic baseline may outperform a sophisticated model on this dataset.

\noindent\textbf{Regularization Term.}
In this set of experiments, we train the models to minimize the loss function of Equation~\eqref{eq:constraint_8} and we report the $1$-step forecasting results in Table~\ref{tab:regularization}.
We also provide some examples of the predictions of the models in Figure~\ref{fig:regularized_mimicking}.
We observe that the proposed regularization term mitigates to some extent the effects of ``mimicking'', however, it does not eliminate it completely.
In most cases, the models trained with the proposed loss function result into a slight increase in MSE compared to the vanilla models, but also into a larger increase in s-MSE.
We also observe that even though the proposed function incurs a very small increase in MSE, it improves the generalization ability of the base models since they achieve higher accuracy in the task of predicting whether the value of the time series will increase or decrease.
The increase in the achieved accuracy of the binary problem  is in some cases significant.
The proposed loss offers LSTM a relative increase of $16.2\%$ in accuracy and Transformer an increase of $6.7\%$ on the Electricity dataset, while TCN's accuracy increases by $9.0\%$ on Sunspots.

\begin{table}[t]
\centering
\begin{minipage}{0.6\linewidth}
    \centering
    \scriptsize
    \def\arraystretch{1.2}
    \caption{$5$-step prediction performance of \\ models on the Electricity dataset. All MSE/ \\ s-MSE results are in scale $(\times 10^{-3})$.}
    \begin{tabular}{|l||cccc|} \hline
    \textbf{Methods} & \textbf{MSE} & \textbf{s-MSE} & \textbf{Acc} & \textbf{s-Acc}  \\ \hline
    Seq2Seq  & 5.495 & 0.522 & 0.375 & 0.605 \\
    Seq2Seq+reg.  & 5.410 & 0.477 & \textbf{0.383} & \underline{0.547} \\ \hline
    LSTM  & 5.534 & 0.502 & 0.380 & 0.646 \\
    LSTM+reg.  & 5.561 & 0.328 & \textbf{0.385} & \underline{0.614} \\ \hline
    TCN  & 5.546 & 0.621 & 0.398 & 0.626 \\
    TCN+reg.  & 5.541 & 0.745 & \textbf{0.401} & \underline{0.610} \\ \hline
    Transf.  & 5.845 & 0.3819 & 0.373 & 0.659 \\
    Transf.+reg.  & 5.346 & 0.3850 & \textbf{0.387} & \underline{0.592} \\ \hline
    \end{tabular}
    \label{tab:multistep}
\end{minipage}
\begin{minipage}{0.35\linewidth}
    \centering
    \scriptsize
    \def\arraystretch{1.2}
    \caption{$1$-step prediction performance of models on the stock price dataset.}
    \begin{tabular}{|l||HHHcc|} \hline
    \textbf{Methods} & \textbf{MAE} & \textbf{MSE} & \textbf{R2} & \textbf{Acc} & \textbf{F1}  \\ \hline
    LSTM  & 3.491 & 4.045 & -0.577 & 0.552 & 0.398 \\ 
    LSTM+reg.  & 3.347 & 4.103 &  -0.672 & \textbf{0.570} & \textbf{0.520} \\ \hline
    TCN & 3.285 & 3.777 & -0.347 & 0.545 & 0.184  \\
    TCN+reg. & 2.803 & 3.413 & -0.146 & \textbf{0.586} & \textbf{0.360} \\ \hline
    \end{tabular}
    \label{tab:stocks}
\end{minipage}
\end{table}

\noindent\textbf{Sensitivity Analysis.}
We next study how the performance of the proposed loss varies as a function of hyperparameter $\lambda$.
We expect the effect of ``mimicking'' to be inversely proportional to $\lambda$.
Figure~\ref{fig:lambdas} illustrates how the performance of the different models on $4$ datasets varies with respect to $\lambda$.
We observe that both MSE and s-MSE increase as the value of $\lambda$ increases.
This is not surprising since the objective of the regularization term is to make s-MSE as large as possible without hurting MSE much.
In most cases, the increase of s-MSE is larger than that of MSE, which is the desired behavior.
In many cases, large values of $\lambda$ result into MSEs that are even greater that that of the baseline (Avg. Window ($n=1$)).
In terms of accuracy, we observe that in most cases, increasing the value of $\lambda$ leads to a slight increase of Acc and a slight decrease of s-Acc.

\noindent\textbf{Multi-step ahead predictions.}
We present in Table~\ref{tab:multistep} results of the multi-step ahead forecasting experiments performed on Electricity.
We employ a sequence-to-sequence model of LSTM encoder and decoder, as well as LSTM, TCN, and Transformer encoders followed by fully connected layers for direct predictions. 
In most cases, when trained to minimize the proposed loss function, the different models achieve slightly larger values of Acc and in some cases significantly smaller values of s-Acc.
In terms of MSE, quite surprisingly in the case of all models except LSTM, MSE decreases when the proposed loss is employed. 

\noindent\textbf{Case Study: Predicting Stock Prices trends.}
We also experiment with a dataset recording high-frequency bids for the TSLA stocks. 
Due to the class imbalance, besides accuracy, we also report F1-scores in Table~\ref{tab:stocks}.
The proposed term leads to slight improvements in accuracy, but significant ones in F1-score.

\section{Conclusion}\label{sec:conclusion}
In this paper, we deal with ``mimicking'' in time series forecasting.
Our results indicate that the proposed regularization term partially mitigates this phenomenon, constituting a first approach towards this research direction.
We plan to further study its properties along with potential improvements in the future. 
Also, investigating the exact conditions under which a model replicates the last observed values of the time series is on our agenda for future work.

\bibliography{icann}
\bibliographystyle{splncs04}

\end{document}